\RequirePackage{fix-cm}
\documentclass[twocolumn]{svjour3}          

\usepackage{geometry}
\usepackage{multirow}
\usepackage{caption}
\usepackage{amssymb}
\usepackage{amsmath}
\usepackage[shortcuts]{extdash}
\usepackage{makecell}
\usepackage{xcolor}
   \definecolor{mygreen}{RGB}{121, 149, 64}
   \definecolor{myblue}{RGB}{0, 112, 192}
   \definecolor{myyellow}{RGB}{191, 144, 0}
   \definecolor{myred}{RGB}{192, 80, 70}

\smartqed  
\usepackage{graphicx}
\usepackage{booktabs}
\usepackage{url}

\usepackage{algpseudocode}
\usepackage{algorithm}

\begin{document}
\title{Spatial-temporal Transformer-guided Diffusion based Data Augmentation for Efficient Skeleton-based Action Recognition}

\author{Yifan Jiang \and Han Chen \and Hanseok Ko}
\institute{
Yifan Jiang, \\\email{yfjiang@korea.ac.kr}
\\ Han Chen, \\\email{jessicachan@korea.ac.kr} 
\\ Hanseok Ko (corresponding author), \\\email{hsko@korea.ac.kr}
\at School of Electrical Engineering, Korea University, Seoul 02841, South Korea
}

\date{Received: date / Accepted: date}
\maketitle

\begin{abstract} 
Recently, skeleton-based human action has become a hot research topic because the compact representation of human skeletons brings new blood to this research domain. As a result, researchers began to notice the importance of using RGB or other sensors to analyze human action by extracting skeleton information. Leveraging the rapid development of deep learning (DL), a significant number of skeleton-based human action approaches have been presented with fine-designed DL structures recently. However, a well-trained DL model always demands high-quality and sufficient data, which is hard to obtain without costing high expenses and human labor. In this paper, we introduce a novel data augmentation method for skeleton-based action recognition tasks, which can effectively generate high-quality and diverse sequential actions. In order to obtain natural and realistic action sequences, we propose denoising diffusion probabilistic models (DDPMs) that can generate a series of synthetic action sequences, and their generation process is precisely guided by a spatial-temporal transformer (ST-Trans). Experimental results show that our method outperforms the state-of-the-art (SOTA) motion generation approaches on different naturality and diversity metrics. It proves that its high-quality synthetic data can also be effectively deployed to existing action recognition models with significant performance improvement. 

\end{abstract}

\section{Introduction}
\label{intro}
Human action recognition is crucial in various video-based visual applications, for instance, video surveillance, video understanding, and human-computer interaction (HCI) \cite{kong2022human,liu2021no,lou2019ar}. Different sensors have been considered, such as RGB frames \cite{fayyaz20213d,duan2020omni,li2021ct}, depth maps \cite{wang2017structured,wang2018depth,sanchez20223dfcnn}, thermal images \cite{imran2020evaluating,mehta2021motion}, and human skeleton \cite{yan2018spatial,shi2019two,liu2020disentangling,cheng2020skeleton,song2022constructing,li2021symbiotic,li20213d}. Among these modalities, the human skeleton is currently gaining growing attention because of its high compactness and robustness.  In practice, the representation of the human skeleton is usually mapped by a time series of 3D coordinate sequences, which can be extracted by pose estimation approaches. Therefore, only the pure skeleton information is included, and it is naturally much more robust to the variation of camera angle, illumination, and background.

While the results of existing works are encouraging, there are always not easy to train a well-performed skeleton-based action recognition model due to data scarcity. In practice, the data scarcity problem is expected due to the high expense of the motion capture and labeling process, but it actually negatively impacts action recognition performance. Therefore, discussing data augmentation for skeleton-based action recognition is meaningful and imperative.

Over the past few years, generative models, represented by generative adversarial networks (GANs) \cite{goodfellow2014generative}, have shown their superiority in different visual tasks, for example, photo-realistic image synthesis \cite{park2019semantic,zhu2020sean}, text-to-image generation \cite{hinz2020semantic,zhu2019dm}, medical image analysis \cite{chen2022unsupervised,jiang2020covid}, image enhancement \cite{kim2020unsupervised,park2019adaptive} and image manipulation \cite{kim2022style,couairon2022flexit,kwak2020cafe}. Nevertheless, GANs' synthetic data suffer from the lack of diversity \cite{nichol2021improved} and low stability when conducting the training process with complex hyper-parameter settings \cite{brock2016neural,brock2018large}. More recently, researchers found that the diffusion models \cite{ho2020denoising} can generate realistic images with high quality. Compared to GANs-based image synthesis approaches, diffusion models are a series of likelihood-based architectures with many advantages: a steady training scheme, better flexibility, and domain adaptive capability \cite{nichol2021improved,dhariwal2021diffusion,nichol2021glide}. Although the above diffusion techniques have recently emerged with encouraging results, generating data containing spatial-temporal information like skeleton sequence leaves much to be desired. Besides, the existing conditional diffusion approach \cite{dhariwal2021diffusion} suffers from low effectiveness because its classifier is pretrained with noisy images using a complex training strategy to enable conditional guidance.

This paper proposes a novel action synthesis algorithm designed to improve skeleton-based action recognition performance in a data-scarcity situation. Specially, we introduce a conditionally generative model that (1) can generate natural action sequence with enough spatial-temporal information, rather than awkward or repeated ones, (2) can be conditioned by specific action categories so that the generation process is controllable, (3) is not constrained to a specific action domain, for instance, actions with sitting or standing poses, (4) does not rely on noisy training data, in other words, can be trained more efficiently and stably.

To achieve the above goals, we design a transformer-guided diffusion model, which consists of two main modules: a visual transformer (ViT) module \cite{mehta2021mobilevit} and a denoising diffusion probabilistic models (DDPM) module \cite{ho2020denoising,nichol2021improved}. Specifically, the pretrained DDPM module cooperates with the pretrained ViT module through a guiding strategy so that DDPM can sample action sequences under the guidance of ViT. Therefore, it can generate a series of augmented action sequences with only an action label provided. To sum up, our contributions are as follows:

\begin{itemize}

\item [(1)] We propose a transformer-guided diffusion approach for improving skeleton-based action recognition performance. It is specially designed and optimized for handling the data scarcity of field-captured action sequences.

\item [(2)] A spatial-temporal transformer is proposed to learn joint position relations on both spatial and temporal levels and precisely guide the diffusion process towards specific action labels.

\item [(3)] We present a novel guiding strategy that enables conditional guidance from the visual transformer to eliminate the dependency on noisy latent. According to our experiments, this strategy is more effective and practical than existing SOTA methods and contributes significantly to diffusion performance.

\end{itemize}

\section{Related works}
\label{Realted works}
\noindent \textbf{Denoising diffusion probabilistic models.} 
Denoising diffusion probabilistic models (DDPMs) consist of a forward diffusion process that gradually inserts noise into inputs and a reverse denoising process that learns to recover data by removing noise. DDPMs have recently been shown to generate high-quality synthetic data, especially images. Many efforts have been made following the invention of DDPMs \cite{ho2020denoising}. Given the limitations of the original DDPMs, some research lands on refining architecture and optimizing sampling strategy. Denoising diffusion implicit models (DDIM) \cite{song2020denoising} accelerate the sampling process by constructing a series of non-Markovian diffusion processes rather than simulating a Markov chain. Later, some scholars introduced critically-damped Langevin diffusion (CLD) \cite{dockhorn2021score} by transferring the successful experience from existing score-based generative models. More recently, DDPMs have been improved in work \cite{nichol2021improved} by optimizing the variational lower-bound to allow DDPMs to achieve better log-likelihoods. On the other hand, some experts lay emphasis on the conditional generation of high-resolution images. As for scalar conditioning, Dhariwal et al. \cite{dhariwal2021diffusion} proposed an Unet \cite{ronneberger2015u} structure that can be integrated into DDPMs and condition the generation process. In order to make DDPMs more controllable, CLIP-based diffusion models \cite{radford2021learning,nichol2021glide,ramesh2022hierarchical} are introduced to leverage the strong visual-language cross-domain representation. Although DDPMs have achieved remarkable results in many domains, they are still rarely seen be utilized for the data augmentation task of skeleton-based action recognition.

\noindent \textbf{Skeleton-based action recognition.} 
With the rapid evolution of pose estimation methods, skeleton-based action recognition approaches are boosted by high-quality skeleton data obtained from advanced pose estimation methods. There are three mainstreams of skeleton-based methods, which are recurrent neural network (RNN) based methods \cite{liu2016spatio,wang2017modeling,lee2017ensemble,shi2019skeleton,li2021memory}, convolutional neural network (CNN) based methods \cite{li2017skeleton,li2018co,caetano2019skelemotion,li2019learning} and GCN based methods \cite{yan2018spatial,li2019actional,shi2019two,liu2020disentangling,song2022constructing,chen2021channel}. In the case of RNN-based methods, they mainly use RNN structure as a long-term temporal learner, which is able to obtain long-range temporal information from input videos. Ref \cite{wang2017modeling} is a Siamese structure that takes both spatial and temporal information at the same time. Liu et al. \cite{liu2016spatio} tried to learn the relationship from one dataset to the other. More recently, \cite{li2021memory} combined the attention mechanism with the RNN model and designed a special temporal attention module that is used for grabbing attention information from the temporal domain of input skeleton sequences. For CNN-based methods, ref \cite{li2017skeleton} brought a new encoding strategy for skeleton data and mapped them into images. Ref \cite{caetano2019skelemotion} also focused on the encoding method, and this work considered both joint motion and temporal information from video together. An end-to-end manner based method \cite{li2018co} was used to utilize different level feature representations. GCN-based methods are the most popular stream of action recognition domain. ST-GCN \cite{yan2018spatial} began to use a graph to represent the spatial and temporal information of skeleton joints. The potential is not limited by predicting current action, and ref \cite{li2019actional,shi2019two} enabled to predict the next action from current skeleton inputs. More recently, MS-G3D \cite{liu2020disentangling} introduced a multi-scale aggregation scheme to disentangle the significance of neighboring joints for better long-range modeling. EfficientGCN \cite{song2022constructing} is a GCN-based method aiming at building a faster and more effective action recognition model by refining network designs. CTR-GCN \cite{chen2021channel} achieved remarkable performance on several popular action recognition datasets leveraging special-designed channel-wise modeling. Despite the encouraging results achieved by the existing skeleton-based methods, few of them consider data scarcity, which is very common in practice.

\noindent \textbf{Conditioned human motion generation.} 
Although generating arbitrary human action is relatively easy and straightforward \cite{ormoneit2005representing,urtasun2007modeling}, its sub-task, the action-conditioned human motion generation, seems much harder and has received less attention recently. Some works have considered transferring different modalities (e.g. text, audio waves, action labels) to human motions. Regarding text-to-motion tasks, some pioneer endeavors have been made on the basis of RNN and advanced language models, Text2Action \cite{ahn2018text2action}, and DVGANs \cite{lin2018human} utilize textual information to generate corresponding motions. As for the audio-to-motion task, in ref \cite{takeuchi2017speech}, an long short-term memory (LSTM) model was proposed to translate audio waves to 3D human gestures. More recently, some efforts related to dance generation have been made \cite{lee2019dancing,li2020learning}. These approaches mainly take music audio as inputs, which condition dance motion generation. Action-conditioned human motion generation is closer to our topic. Action2Motion \cite{guo2020action2motion} is a variational auto-encoder (VAE) based model, designed for generating diverse human actions. And ACTOR \cite{petrovich2021action} presented a novel transformer-based VAE to solve the variable-length motion generation problem. Although the proposed method is also an action-conditioned approach, we try to tackle the motion generation problem in another way: leveraging modern image synthesis techniques to synthesize realistic and natural human motion.

\begin{figure}[!ht]
\centering
\includegraphics[width=8cm]{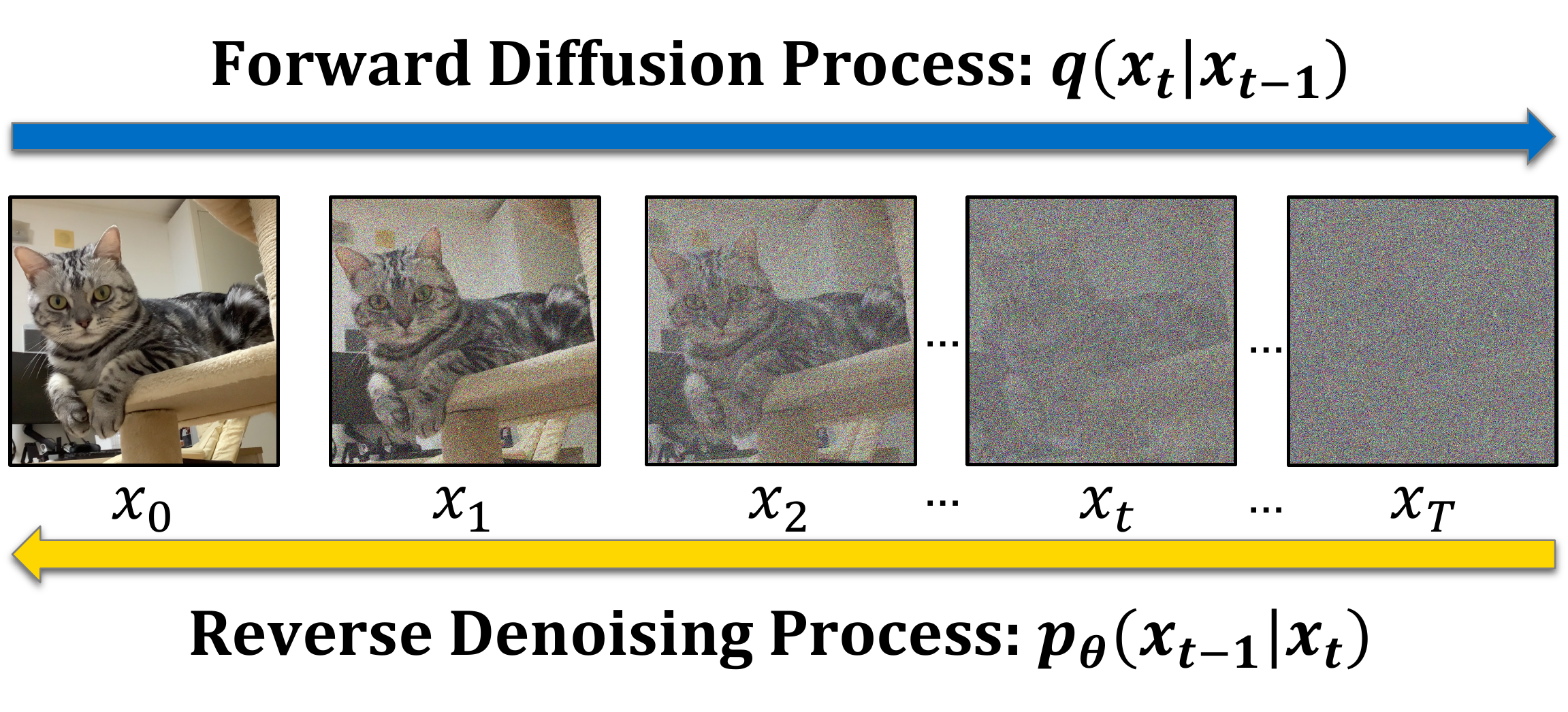}
\caption{Workflow of Denoising Diffusion Probabilistic Models (DDPMs). The forward and reverse processes are denoted as blue and yellow arrows, respectively.}
\label{fig1} 
\end{figure}

\begin{figure*}[!ht]
\centering
\includegraphics[width=13cm]{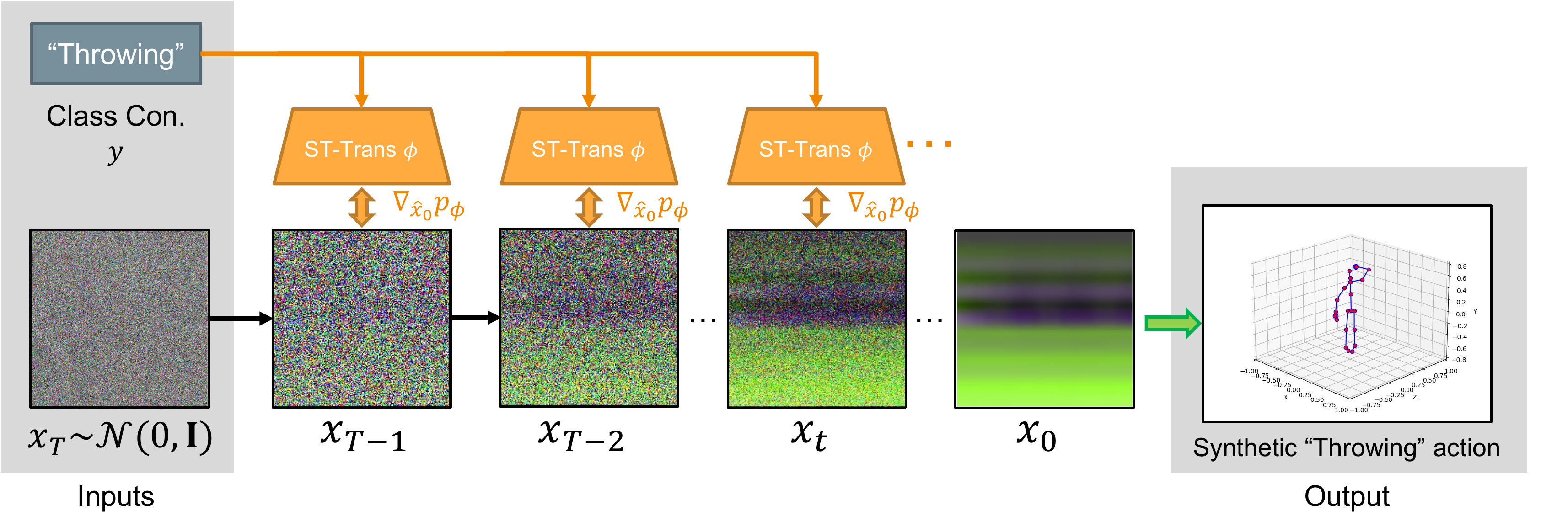}
\caption{Overview of the proposed method. Orange arrows show the transformer-guided process, black arrows indicate the reverse denoising process, and green arrow shows the translation process from the skeleton image representation (we do not show the skeleton image representation with zero paddings for a nice presentation) to the 3D joint coordinate representation. Moreover, we denote spatial-temporal transformer as ST-Trans in short.}
\label{fig2} 
\end{figure*}

\section{DDPMs preliminaries}
\label{DDPMs}

In this subsection, we reviewed the Denoising Diffusion Probabilistic Models (DDPMs) structures and formulations by following the notations in ref \cite{nichol2021improved}. The DDPMs workflow is shown in Figure \ref{fig2}, and we can separate DDPMs into two processes: a forward diffusion process that adds noise into an input $x_0$ gradually by the timestep $t$ to obtain an isotropic Gaussian noise sample $x_T$. Then a reverse denoising process is conducted, and samples intermediate latent $x_{T-1}, x_{T-2}...$ from $x_T$ step by step towards a clean sample $x_0$. 

During the forward process, we start from an input $x_0$, which obeys Gaussian distribution $q(x_0)$. Then, we derive a series of intermediate latent $x_1,...,x_t$ by noising them with Gaussian noise with variance $\beta_t$ at each timestep $t$:

\begin{equation}
\begin{aligned}
q(x_t|x_{t-1})=\mathcal{N}(x_t;\sqrt{1-\beta_t}x_{t-1},\beta_t\textbf{I})
\\ jointly \quad q(x_{1:T}|x_0) = \prod_{t=1}^Tq(x_t|x_{t-1})
\end{aligned}
\label{eq1}
\end{equation}

To sample intermediate latent $x_t$ faster, we define $\bar{\alpha}_t=\prod_{s=1}^t(1-\beta_s)$, therefore, we can acquire a diffusion kernel, which can be used to sample $x_t$ as follow:

\begin{equation}
\begin{aligned}
q(x_t|x_0)=\mathcal{N}(x_t;\sqrt{\bar{\alpha}_t}x_0,(1-\bar{\alpha}_t)\textbf{I})
\\ Sampling \quad x_t=\sqrt{\bar{\alpha}_t}x_0+\sqrt{(1-\bar{\alpha}_t)}\epsilon
\end{aligned}
\label{eq2}
\end{equation}
where $\epsilon\sim\mathcal{N}(0,\textbf{I})$.

In order to re-sample a new data point from the distribution $q(x_0)$, we need to follow the reverse denoising process, which starts from sampling data from the distribution $p(x_T)=\mathcal{N}(x_T;0,\textbf{I})$, then, continuously keeps sampling the posteriors $q(x_{t-1}|x_t)$. Since $q(x_{t-1}|x_t)$ is intractable, we opt for a learnable variational autoencoder (VAE) $p_\theta$ to approximate the posteriors by predicting the mean $\mu_\theta(x_t,t)$ and variance $\sigma_t^2$ of $x_{t-1}$ given input $x_t$. Therefore, intermediate latent $x_{t-1}$ and a new data point $x_0$ is able to be sampled from the distribution as follows:

\begin{equation}
\begin{aligned}
p_\theta(x_{t-1}|x_t)=\mathcal{N}(x_{t-1};\mu_\theta(x_t,t), \sigma_t^2\textbf{I})
\\ p_\theta(x_{0:T})=p(x_T)\prod_{t=1}^Tp_\theta(x_{t-1}|x_t)
\end{aligned}
\label{eq3}
\end{equation}

In practice, the strategy of obtaining the mean and variance is tricky. As for the mean $\mu_\theta(x_t,t)$, Ho et al. \cite{ho2020denoising} introduced a noise-prediction network to obtain the mean by predicting the noise $\epsilon_\theta(x_t,t)$. So we can update Equation \ref{eq2} as below:

\begin{equation}
\begin{aligned}
\mu_\theta(x_t,t)=\frac{1}{\sqrt{1-\beta_t}}(x_t-\frac{\beta_t}{\sqrt{1-\bar{\alpha}_t}}\epsilon_\theta(x_t,t))
\end{aligned}
\label{eq4}
\end{equation}

In terms of variance $\sigma_t^2$, Ho et al. \cite{ho2020denoising} kept $\sigma_t^2=\beta_t$. However, recent works \cite{nichol2021improved,bao2022analytic} found that a parameterized $\sigma_t^2$ by minimizing the variational bound leads to faster convergence and a more stable training process.

\section{Spatial-temporal Transformer-guided Diffusion-based Data Augmentation for Efficient Skeleton-based Action Recognition}
\label{Method}

We introduce the proposed method in this section with details. In Figure \ref{fig3}, we demonstrate the overview of the proposed method. Given a class condition $y$, we aim to generate natural action sequences using a pretrained diffusion model $\theta$ under the guidance of a pretrained spatial-temporal transformer $\phi$. We start from a pretrained diffusion model with two inputs: a class condition $y$ and a noise map $x_t$ sampled from Gaussian distribution $\mathcal{N}(0,\textbf{I})$. Next, a series of intermediate latent $x_{t-1}...x_1$ are sampled step by step. At each sampling step, the process is guided by the pretrained transformer $p_\phi(y|\hat{x}_0)$ using its gradient $\nabla_{\hat{x}_0}p_\phi$ after acquiring the clean estimation $\hat{x}_0$ of the noisy latent $x_t$. So that this guiding mechanism leads the sampling process gradually toward the class condition $y$. The final output is a synthetic skeleton image, and this skeleton image representation is then translated to 3D joint coordinates and can be easily used by action recognition methods.

\subsection{Transformer-guided Diffusion}

\begin{algorithm}
\caption{Transformer-guided diffusion: given a pretrained diffusion model $(\mu_\theta(x_t), \sigma_\theta(x_t))$ and a pretrained spatial-temporal transformer $p_\phi(y|\hat{x}_0)$.}\label{al1}
\textbf{Input:} target class label $y$\\
\textbf{Output:} synthetic action sequence in skeleton image representation $x_0$ according to target class label $y$ \\
$x_T\gets$ sample from $\mathcal{N}(0,\textbf{I})$
\begin{algorithmic}[1]
\For{\textbf{all} $t$ from $T$ to 1}
    \State $\mu,\sigma\gets\mu_\theta(x_t), \sigma_\theta(x_t)$
    \State $\hat{x}_0\gets\frac{1}{\sqrt{\bar{\alpha}_t}}[x_t-\sqrt{1-\bar{\alpha}_t}\epsilon_\theta(x_t,t)]$
    \State $\nabla_{\hat{x}_0}p_\phi\gets\nabla_{\hat{x}_0}\mathcal{L}_C(\hat{x}_0, y)$
    \State $x_{t-1}\gets\mathcal{N}(\mu+\sigma\nabla_{\hat{x}_0}p_\phi,\sigma)$
\EndFor
\end{algorithmic}
\textbf{return} $x_0$
\end{algorithm}

Dhariwal et al. \cite{dhariwal2021diffusion} proposed a conditional diffusion method, which can leverage a classifier pretrained on noisy images to guide the sampling process toward a class condition. Nevertheless, in order to learn the representation of noisy data, its classifier design and training process is complex and ineffective. To tackle this issue, we propose a novel guiding strategy, which directly estimates a clean image $\hat{x}_0$ from the intermediate latent $x_t$ and uses it later to obtain the gradient of the transformer. Recall the DDPMs preliminaries, we can estimate the noise in each timestep $\epsilon_\theta(x_t,t)$, which was added to $x_0$ to acquire $x_t$. Naturally, the clean image $x_0$ can be derive from $\epsilon_\theta(x_t,t)$ through Equation \ref{eq2}:

\begin{equation}
\begin{aligned}
\hat{x}_0=\frac{1}{\sqrt{\bar{\alpha}_t}}[x_t-\sqrt{1-\bar{\alpha}_t}\epsilon_\theta(x_t,t)]
\end{aligned}
\label{eq5}
\end{equation}

Finally, we define a loss to evaluate the similarity between synthetic and real action sequences. Specifically, we propose to use a simple but effective cross-entropy loss $\mathcal{L}_C(\hat{x}_0, y)$ to evaluate the difference between synthetic and target data distributions. To clearly depict the whole transformer-guided diffusion process, we summarize the proposed method in Algorithm \ref{al1}.

\subsection{Spatial-temporal transformer}

In this subsection, we discuss the proposed spatial-temporal transformer in detail. Since the skeleton image representation is in a tiny size and contains rich spatial-temporal information (temporal joint position change). Therefore, it is natural to opt for a visual transformer to deal with these data because ViT has a superior attention mechanism, which can effectively learn relations among patches from the original image.

In this work, we use MobileViT \cite{mehta2021mobilevit} as our backbone and adjust the network design to our task. Since an action sequence is represented as a tiny size image, we refine the network structure to adjust to the input and modify the internal  MobileNetv2 and MobileViT blocks. This improvement allows us to train the proposed spatial-temporal transformer faster and improve the guidance performance for small images.

Experimental results suggest that the proposed spatial-temporal transformer outperforms other state-of-the-art CNNs-based classifiers when guiding a diffusion process in different metrics by leveraging its indigenous attention mechanism. Additionally, its lightweight and compact design enables it to perform better than the other classifiers in the tiny-size skeleton image representation.

\begin{table*}[]
	\centering
	\caption{Naturality and diversity evaluation on the HumanAct12 dataset. Acc. is action recognition accuracy, O. Div. is overall diversity, and PA. Div. is per-action diversity. (The best evaluation score is marked in bold. $\uparrow$ means a higher number is better, $\downarrow$ indicates a lower number is better, and $\rightarrow$ means the number closer to Real actions is better. $\pm$ indicates 95\% confidence interval.)}
 
 \renewcommand{\arraystretch}{1.5}
	\begin{tabular}{c c c c c}
		\toprule
            
            Methods  & FID ($\downarrow$) & Acc. ($\uparrow$) & O. Div. ($\rightarrow$) & PA. Div. ($\rightarrow$)  \\
            
		\midrule
            Real actions & $0.09^{\pm 0.01}$& $0.96^{\pm 0.01}$ & $6.74^{\pm 0.03}$ & $2.55^{\pm 0.02}$ \\ 

            \midrule

            CondGRU \cite{shlizerman2018audio,guo2020action2motion} &  $39.92^{\pm 0.13}$& $0.06^{\pm 0.03}$ & $2.05^{\pm 0.05}$ & $2.18^{\pm 0.02}$ \\

            Two-stage GAN \cite{cai2018deep,guo2020action2motion}  & $12.08^{\pm 0.11}$& $0.45^{\pm 0.01}$ & $5.35^{\pm 0.06}$ & $2.21^{\pm 0.03}$ \\

            Act-MoCoGAN \cite{tulyakov2018mocogan,guo2020action2motion}  & $5.73^{\pm 0.18}$& $0.77^{\pm 0.01}$ & $\textbf{6.84}^{\pm 0.04}$ & $1.26^{\pm 0.02}$ \\
            
            Action2Motion \cite{guo2020action2motion} & $2.66^{\pm 0.09}$& $0.91^{\pm 0.01}$ & $6.98^{\pm 0.03}$ & $2.88^{\pm 0.01}$ \\
            
            ACTOR \cite{petrovich2021action} & $0.24^{\pm 0.03}$& $0.93^{\pm 0.01}$ & $6.62^{\pm 0.05}$ & $2.49^{\pm 0.03}$ \\

            \midrule

            Ours & $\textbf{0.12}^{\pm 0.01}$ & $\textbf{0.95}^{\pm 0.01}$ & $6.88^{\pm 0.02}$ & $\textbf{2.50}^{\pm 0.02}$ \\

    	\bottomrule
	\end{tabular}
\label{table1}
\end{table*}

\section{Experiments}
\label{Experiments}
\subsection{Dataset and experimental settings}
\subsubsection{Dataset}

\noindent \textbf{NTU RGB+D} \cite{shahroudy2016ntu,liu2020ntu} 
NTU RGB+D dataset is a large-scale video dataset for the action recognition task. It contains a total of 120 different action categories and 114,480 video clips ranging from daily actions to two-person interactions. To evaluate data augmentation performance for the action recognition task, we evaluate the proposed method on the full-scale NTU RGB+D dataset. In order to compare to other SOTA action recognition original performances, we use the original pose annotations, which are captured by Kinect. There are two benchmarks suggested: 1) cross-subject benchmark separates video clips into a training set (63,026) and an evaluation set (50,922) by subject characteristics. 2) cross-setup benchmark separates video clips into a training set (54,471) and evaluation set (59,477) by scenario setups. We take 20\% of the training set in subject or setup levels to pretrain the diffusion and ST-Trans models and leave the rest of the training set for the training procedure of action recognition approaches in the quantitative experiment. Note that we remove the video clips which are mutual actions or overlong/-short or too noisy in the original dataset.

\noindent \textbf{HumanAct12}
Similar to NTU RGB+D, we follow the experimental settings of the other two SOTA methods \cite{guo2020action2motion,petrovich2021action} to utilize the HumanAct12 dataset to evaluate the naturality and diversity of the proposed method. HumanAct12 dataset is an adjusted version of the PHSPD dataset \cite{zou20203d,zou2020polarization}. All 1,191 video clips are reorganized into 12 action categories, and the corresponding SMPL parameters are also provided. 

\subsubsection{Evaluation metrics}

To fully evaluate the proposed method's performances and compare them to other SOTA action-conditioned motion generation methods, and evaluate the data augmentation performance properly, we propose two kinds of evaluation metrics in this paper:

\subsubsection{Evaluation metrics for naturality and diversity} 

We follow \cite{guo2020action2motion} and \cite{petrovich2021action} to measure the naturality and diversity of synthetic data. Frechet Inception Distance (FID), action recognition accuracy, overall diversity, and per-action diversity, a total of four metrics, are considered in the naturality and diversity experiments. To be specific, FID \cite{heusel2017gans} is a prevalent metric to evaluate the similarity between synthetic and real data. A lower FID score indicates that the synthetic data is closer to the real data. Additionally, we apply the same pretrained RNN-based action recognition model in \cite{guo2020action2motion} and \cite{petrovich2021action} on a set of synthetic action sequences and report the action recognition accuracy. A higher accuracy indicates that the synthetic data distribution is more similar to the real one. As for the overall diversity metric, we extract the features from a set of synthetic and real data using the above pretrained RNN model, then compute the L2 distance between each synthetic-real feature pair. Finally, we utilize the per-action diversity from an L2 distance between each synthetic-real feature pair above but at the class level. 

\subsubsection{Evaluation metrics of data augmentation task for action recognition} 

Most SOTA action recognition methods report their cross-subject and cross-setup accuracy on NTU RGB+D 120 dataset. In order to make our results comparable, we follow these two metrics when evaluating the data augmentation performance for the action recognition task.

\subsection{Implementation details}

\subsubsection{Skeleton image representation} 

Since image synthesis in this work is just an intermediate stage, the image representation should be able to be translated back to the joint coordinates losslessly. We follow Du et al. \cite{du2015skeleton} to encode the action sequence into a matrix, which has the size of $J\times T \times 3$, where $J$ is the number of joints, $T$ indicates the length of the corresponding action sequence and 3 is the 3D coordinates of each joint. To avoid resizing operations that may change the pixel value of the skeleton image representation, we keep $J=T$ to acquire a square skeleton image representation. In practice, this image representation will be centrally interpolated with zero paddings into a $32 \times 32$ image as the model input.

\subsubsection{Experimental details} 

The diffusion model and the spatial-temporal transformer are pretrained using  HumanAct12 and NTU RGB+D following the split of cross-sub or cross-subject. The initial learning rate is set as $1e-4$. Both models are trained using AdamW \cite{loshchilov2017decoupled} with $\beta_1=0.9$ and $\beta_2=0.999$ for 500K iterations. We keep the batch size of each model as 1024. In particular, some hyper-parameters are important to diffusion models. We reference Nicho et al. \cite{nichol2021improved} to design the diffusion network, which has 128 base channels and three residual blocks per resolution. And we follow the geometric losses introduced by \cite{shi2020motionet,petrovich2021action} to train the proposed model. Furthermore, we set the number of diffusion steps as 1,000 and used a cosine noise schedule during the training stage. Additionally, we generate synthetic datasets 20 times randomly using different random seeds to report the average with a confidence interval of 95\%. As for the experimental environment, all the experiments are conducted through PyTorch in an Ubuntu 18.04 platform with Intel 9700K CPU and two Nvidia RTX Titans. 

Regarding action recognition models, we follow the experimental settings of the authors. Please refer to the original papers for details.

Note that since we use a different representation of action sequences and re-train all the methods we want to compare, the experimental results of SOTA competitors in this paper may differ from the original publications. 

\begin{table*}[]
	\centering
	\caption{Replacement data augmentation experiment on the NTU RGB+D 120 dataset. 0\%-50\% means replacing real training data using synthetic data from the proposed method (or Action2Motion and ACTOR) with different proportions. Results are reported as $(\begin{matrix}Cross-Subject \\ Cross-Setup\end{matrix})$ accuracy. (The best evaluation score is marked in bold. $\pm$ indicates 95\% confidence interval.)}
 
 \resizebox{\textwidth}{18mm}{
 \renewcommand{\arraystretch}{2}
	\begin{tabular}{c c c c c c c | c c}
		\toprule
            Methods & 0\% & 10\% & 20\% & 30\% & 40\% & 50\% & 50\% (A2M) & 50\% (ACTOR)  \\
            
		\midrule

            MS-G3D \cite{liu2020disentangling} & 
            \makecell{$71.4^{\pm 0.0}$ \\ $72.0^{\pm 0.0}$} & \makecell{$70.5^{\pm 0.2}$ \\ $71.2^{\pm 0.2}$} & \makecell{$74.6^{\pm 0.1}$ \\ $\textbf{76.9}^{\pm 0.2}$} & \makecell{$\textbf{75.4}^{\pm 0.1}$ \\ $\textbf{76.9}^{\pm 0.2}$} & 
            \makecell{$73.3^{\pm 0.2}$ \\ $73.9^{\pm 0.2}$} & \makecell{$70.9^{\pm 0.4}$ \\ $71.1^{\pm 0.3}$} &
            \makecell{$62.7^{\pm 0.2}$ \\ $64.2^{\pm 0.3}$} & \makecell{$66.8^{\pm 0.3}$ \\ $67.1^{\pm 0.3}$} \\

            EfficientGCN-B4 \cite{song2022constructing} & 
            \makecell{$72.2^{\pm 0.0}$ \\ $72.6^{\pm 0.0}$} & \makecell{$72.1^{\pm 0.1}$ \\ $72.3^{\pm 0.2}$} & \makecell{$\textbf{75.8}^{\pm 0.3}$ \\ $\textbf{77.2}^{\pm 0.1}$} & \makecell{$75.6^{\pm 0.2}$ \\ $77.0^{\pm 0.1}$} & 
            \makecell{$74.1^{\pm 0.1}$ \\ $74.8^{\pm 0.2}$} & \makecell{$72.0^{\pm 0.3}$ \\ $73.1^{\pm 0.2}$} &
            \makecell{$63.6^{\pm 0.3}$ \\ $64.9^{\pm 0.3}$} & \makecell{$67.0^{\pm 0.2}$ \\ $67.8^{\pm 0.4}$} \\

            CTR-GCN \cite{chen2021channel} & 
            \makecell{$72.5^{\pm 0.0}$ \\ $73.4^{\pm 0.0}$} & \makecell{$73.2^{\pm 0.1}$ \\ $73.9^{\pm 0.1}$} & \makecell{$76.4^{\pm 0.2}$ \\ $77.6^{\pm 0.1}$} & \makecell{$\textbf{76.9}^{\pm 0.1}$ \\ $77.4^{\pm 0.1}$} & 
            \makecell{$76.1^{\pm 0.2}$ \\ $\textbf{77.7}^{\pm 0.1}$} & \makecell{$72.5^{\pm 0.3}$ \\ $73.5^{\pm 0.3}$} &
            \makecell{$63.9^{\pm 0.4}$ \\ $64.8^{\pm 0.2}$} & \makecell{$69.3^{\pm 0.3}$ \\ $70.0^{\pm 0.4}$} \\
      
    	\bottomrule
	\end{tabular}
 }
\label{table2}
\end{table*}

\begin{table*}[]
	\centering
	\caption{Incremental data augmentation experiment on the NTU RGB+D 120 dataset. 0\%-50\% means adding extra synthetic data from the proposed method (or Action2Motion and ACTOR) into real training data with different proportions. Results are reported as $(\begin{matrix}Cross-Subject \\ Cross-Setup\end{matrix})$ accuracy. (The best evaluation score is marked in bold. $\pm$ indicates 95\% confidence interval.)}
 
 \resizebox{\textwidth}{18mm}{
 \renewcommand{\arraystretch}{2}
	\begin{tabular}{c c c c c c c | c c}
		\toprule
            Methods & 0\% & 10\% & 20\% & 30\% & 40\% & 50\% & 50\% (A2M) & 50\% (ACTOR)  \\
            
		\midrule

            MS-G3D \cite{liu2020disentangling} & 
            \makecell{$71.4^{\pm 0.0}$ \\ $72.0^{\pm 0.0}$} & \makecell{$71.6^{\pm 0.1}$ \\ $72.3^{\pm 0.1}$} & \makecell{$72.5^{\pm 0.1}$ \\ $73.0^{\pm 0.2}$} & \makecell{$74.2^{\pm 0.2}$ \\ $74.9^{\pm 0.3}$} & 
            \makecell{$\textbf{75.6}^{\pm 0.1}$ \\ $76.0^{\pm 0.3}$} & \makecell{$75.5^{\pm 0.1}$ \\ $\textbf{76.2}^{\pm 0.1}$} &
            \makecell{$65.6^{\pm 0.3}$ \\ $65.8^{\pm 0.4}$} & \makecell{$69.9^{\pm 0.3}$ \\ $70.4^{\pm 0.2}$} \\

            EfficientGCN-B4 \cite{song2022constructing} & 
            \makecell{$72.2^{\pm 0.0}$ \\ $72.6^{\pm 0.0}$} & \makecell{$73.0^{\pm 0.1}$ \\ $73.6^{\pm 0.1}$} & \makecell{$73.3^{\pm 0.1}$ \\ $73.5^{\pm 0.1}$} & \makecell{$\textbf{76.8}^{\pm 0.2}$ \\ $\textbf{77.9}^{\pm 0.2}$} & 
            \makecell{$76.7^{\pm 0.3}$ \\ $77.3^{\pm 0.2}$} & \makecell{$76.5^{\pm 0.1}$ \\ $77.0^{\pm 0.3}$} &
            \makecell{$67.0^{\pm 0.3}$ \\ $67.8^{\pm 0.4}$} & \makecell{$71.8^{\pm 0.2}$ \\ $72.5^{\pm 0.4}$} \\

            CTR-GCN \cite{chen2021channel} & 
            \makecell{$72.5^{\pm 0.0}$ \\ $73.4^{\pm 0.0}$} & \makecell{$73.0^{\pm 0.2}$ \\ $73.7^{\pm 0.1}$} & \makecell{$74.6^{\pm 0.1}$ \\ $75.3^{\pm 0.1}$} & \makecell{$77.4^{\pm 0.3}$ \\ $78.9^{\pm 0.2}$} & 
            \makecell{$\textbf{77.6}^{\pm 0.4}$ \\ $\textbf{79.4}^{\pm 0.2}$} & \makecell{$77.4^{\pm 0.2}$ \\ $79.1^{\pm 0.3}$} &
            \makecell{$67.9^{\pm 0.3}$ \\ $69.0^{\pm 0.3}$} & \makecell{$72.2^{\pm 0.4}$ \\ $72.6^{\pm 0.2}$} \\
      
    	\bottomrule
	\end{tabular}
 }
\label{table3}
\end{table*}

\subsection{Quantitative Results}

\subsubsection{Naturality and diversity evaluation}

In Table \ref{table1}, we summarise the naturality and diversity experimental results on the HumanAct12 datasets. The proposed method is compared with several SOTA action-conditioned motion generation methods, which are introduced as follows: 

\begin{itemize}

\item \textbf{CondGRU} is originally a RNN based audio-to-motion approach \cite{shlizerman2018audio}, which modified by \cite{guo2020action2motion} adjusting the network to receive the condition vector and pose vector. 

\item \textbf{Two-stage GAN} \cite{cai2018deep} uses a motion generator to create a noise vector, which can be used to generate 2D motion sequences. Action2Motion's authors \cite{guo2020action2motion} managed to enable the Two-stage GAN to work for 3D motions. 

\item \textbf{Act-MoCoGAN} \cite{tulyakov2018mocogan} is a video generation method synthesizing realistic video clips using noise vectors and certain content as inputs. Guo et al. \cite{guo2020action2motion} updated it with different discriminators to be suitable for motion generation tasks.

\item \textbf{Action2Motion} \cite{guo2020action2motion} is a gated recurrent unit (GRU) based VAE structure, which can generate natural motion sequences by action conditions at the frame level.

\item \textbf{ACTOR} \cite{petrovich2021action} is also a VAE based approach, but relies on a transformer architecture to perform encoding and decoding operations.

\end{itemize}

We can observe the proposed method is able to outperform not only old-fashioned GAN-based methods but also two recent VAE-based approaches in different naturality and diversity metrics. The significant improvements come from the novel diffusion structure, which (1) can leverage its strong capability on image synthesis tasks to generate noise-less and diverse images, and (2) is guided by a spatial-temporal transformer generating motion sequences stably and precisely. Therefore, we can obtain more natural motion sequences with different action categories through the proposed method.

Although the proposed method is not able to achieve the best result on some metrics, it shows its higher robustness and lower fluctuation on these metrics compared to other SOTA approaches.

\subsubsection{Data augmentation evaluation}

In this subsection, we discuss the experiment of data augmentation for action recognition. The experiment is divided into two parts: (1) Replacing real training data using different proportions of synthetic data; (2) Adding extra synthetic data into training data with different proportions.

\noindent \textbf{Replacement experiment.} We present the experimental results of the replacement experiment in Table \ref{table2}. We conduct the experiment by replacing the training set of three SOTA action recognition methods \cite{liu2020disentangling,song2022constructing,chen2021channel} with different proportions of synthetic data. It is easy to observe that the action recognition accuracy is increasing along with replacing a larger amount of real data. The performance peaks at 20\%-30\% replacement ratio and decreases when the replacement ratio reaches 50\%. The experimental results suggest that the synthetic data created by the proposed method is natural and diverse enough to bring improvements to recent SOTA action recognition models. Moreover, we also involve the synthetic data from two motion generation competitors, and the 50\% replacement evaluation results tell that the proposed method can generate more realistic and diverse action sequences, which can be more useful in a downstream action recognition task.

\noindent \textbf{Incremental experiment.} We summarize the experimental results of the incremental experiment in Table \ref{table3}. Similar to the replacement experiment, we apply a mixed training set of real and synthetic data on three SOTA action recognition methods. Rather than replacing real data with synthetic ones, we add extra synthetic data generated by the proposed method or the other two competitors. The experimental results suggest that action recognition performance increases significantly when we bring more synthetic action sequences into the training set and the performance peaks when 40\% synthetic data is added. Furthermore, compared to the two competitors, the synthetic data generated by the proposed method can offer more performance gain to the SOTA action recognition models and make the performance more stable.

\subsection{Qualitative Results}

\begin{figure}[!ht]
\centering
\includegraphics[width=8cm]{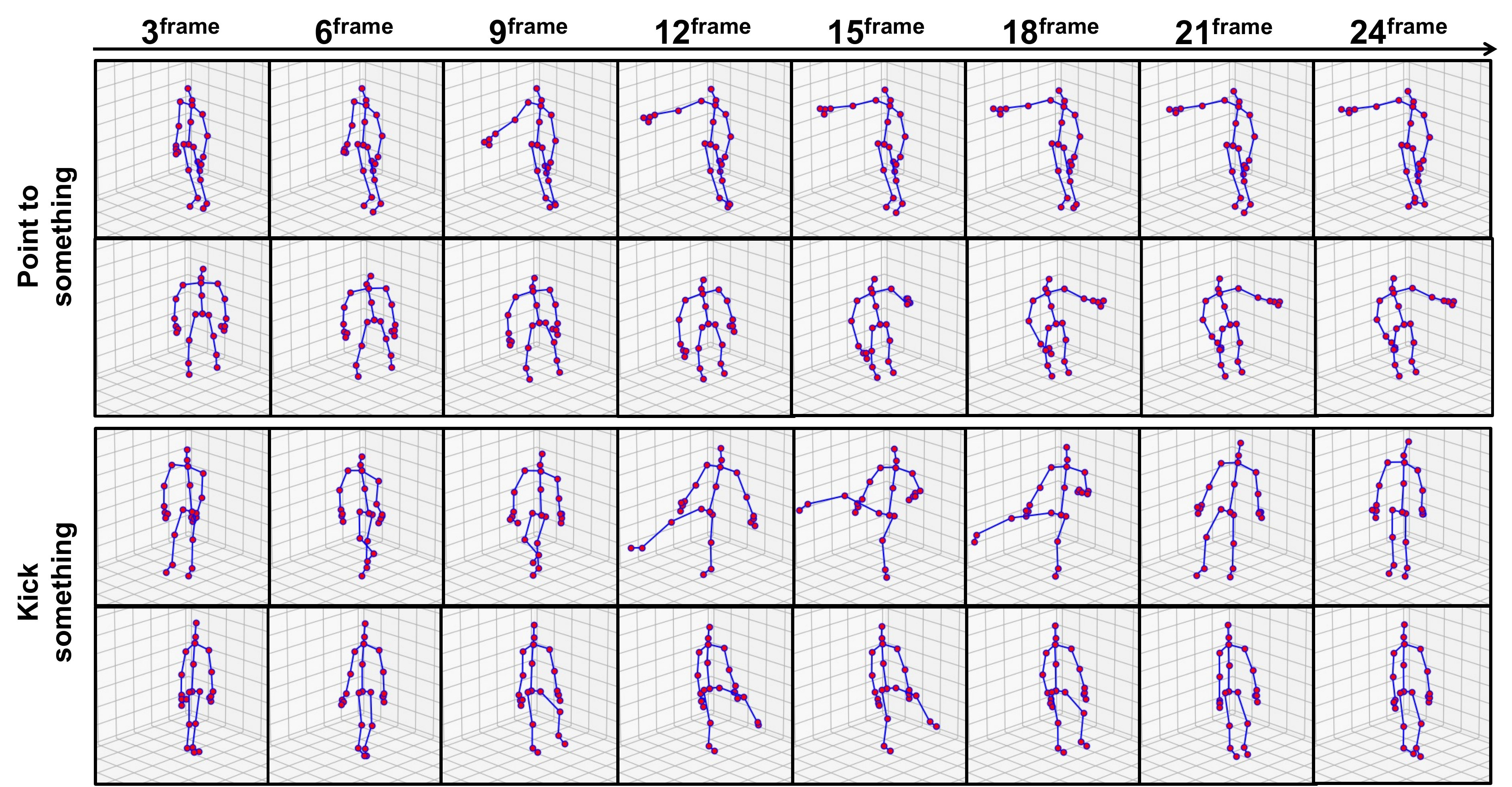}
\caption{Examples of synthetic 'point to something' (top) and 'kick something' (bottom) action sequences.}
\label{fig3} 
\end{figure}

Figure \ref{fig3} depicts some examples of synthetic action sequences with two different labels, which are generated from the proposed method. Eight samples are selected from a single action sequence.

As for the examples of 'point to something' action sequences, the proposed method is able to generate a natural and smooth action sequence, which also contains almost no noise. In terms of the examples of 'kick something' action sequences, the proposed method can generate a realistic kicking action sequence with diverse and normal pose representations.

\subsection{Ablation Studies}

In this subsection, we further discuss the components of the proposed method and their contributions to the performance. 

\begin{table}[htbp] 
 \centering
 \caption{Ablation studies of different guiding models and hyper-parameters. 50\% replacement experiment settings are used. CS: Cross-Subject accuracy, CP: Cross-Setup accuracy. ($\pm$ indicates 95\% confidence interval.)} 
 \begin{tabular}{cccc} 
  \toprule 
  Method & FID ($\downarrow$) & CS ($\uparrow$) & CP ($\uparrow$) \\ 
  \midrule 
  \makecell{Ours (ST-Trans \\ + 1000 steps \\ + Cosine)} & $0.39^{\pm 0.01}$ & $72.5^{\pm 0.3}$ & $73.5^{\pm 0.3}$ \\
  \midrule 
 w/o Guidance & $39.83^{\pm 0.26}$ & - & - \\
 CNNs \cite{tan2019efficientnet} & $3.49^{\pm 0.02}$ & $61.8^{\pm 0.5}$ & $63.4^{\pm 0.3}$ \\
 Unet \cite{dhariwal2021diffusion} & $1.66^{\pm 0.03}$ & $70.7^{\pm 0.3}$ & $71.2^{\pm 0.4}$ \\
 CLIP \cite{radford2021learning} & $2.92^{\pm 0.04}$ & $68.2^{\pm 0.3}$ & $68.8^{\pm 0.3}$ \\
 \midrule 
 100 steps & $12.50^{\pm 0.13}$ & $66.4^{\pm 0.2}$ & $66.9^{\pm 0.1}$ \\
 500 steps & $1.46^{\pm 0.01}$ & $69.2^{\pm 0.2}$ & $70.1^{\pm 0.3}$ \\
 2500 steps & $0.41^{\pm 0.01}$ & $72.5^{\pm 0.4}$ & $73.3^{\pm 0.3}$ \\
 \midrule 
 Linear & $0.40^{\pm 0.01}$ & $72.6^{\pm 0.5}$ & $72.9^{\pm 0.4}$ \\
  \bottomrule 
 \end{tabular} 
\label{table4}
\end{table}

Table \ref{table4} summarizes the ablation studies about different guiding models and hyper-parameters. From the second part of Table \ref{table4}, we compare the FID and action recognition accuracy, which different guiding models report. EfficientNet \cite{tan2019efficientnet}, a Unet-based classifier \cite{dhariwal2021diffusion} and a fine-tuned CLIP \cite{radford2021learning} are applied as guiding classifiers. The experiment results show that the proposed spatial-temporal transformer outperforms other competitors due to its high capability of dealing with action sequences containing rich spatial-temporal information. In addition, we found that the number of diffusion steps less than 1,000 (e.g. 250 and 500) is too small for acquiring high-quality action sequences, and the number of more than 1,000 (e.g. 2,500) gains no obvious performance increase. Finally, we compared the linear and cosine noise schedules and found that the cosine noise schedule is better for skeleton image generation because the performance in different metrics is more stable.

\begin{table}[htbp] 
 \centering
 \caption{Ablation studies of the effectiveness of various guiding strategies. ($\pm$ indicates 95\% confidence interval.)} 
 \begin{tabular}{ccc} 
  \toprule 
  Method & Model size ($\downarrow$) & \makecell{Convergence \\ iterations ($\downarrow$)} \\ 
  \midrule 
  \makecell{Ours (ST-Trans \\ + Clean data)} & 196M & $463K^{\pm 2.6K}$  \\
  \midrule 
  \makecell{Unet \cite{dhariwal2021diffusion} + \\ Noisy data} & 223M & $533K^{\pm 2.0K}$  \\
  \makecell{CNNs \cite{tan2019efficientnet} + \\ Clean data} & 219M & $480K^{\pm 3.1K}$  \\
  \makecell{ViT \cite{dosovitskiy2020image} + \\ Clean data} & 240M & $561K^{\pm 4.9K}$  \\
  \makecell{CLIP \cite{radford2021learning} + \\ Clean data} & 261M & $249K^{\pm 3.3K}$  \\
  \bottomrule 
 \end{tabular} 
\label{table5}
\end{table}

In Table \ref{table5}, the ablation studies about the effectiveness of various guiding strategies are summarized. The experimental results suggest that the proposed method has a smaller model size but a faster convergence speed. A looser dependency on clean training data enables an easier data preparation process and a more straightforward guided-diffusion model design.

\section{Conclusions and Future Studies}
\label{Conclusion}
In this paper, we introduced a novel spatial-temporal transformer-guided diffusion model for action recognition data augmentation tasks. The proposed method takes an action label as an input, then generates high-quality action sequences with the corresponding target labels under the guidance of a spatial-temporal transformer. During the generation process, the proposed spatial-temporal transformer classifies clean intermediate latent generated step-by-step by sampling from a Gaussian distribution. With the experimental results on the naturality and diversity evaluation and the data augmentation evaluation, the proposed method showed the superior capability of synthesizing high-quality action sequences compared to the existing SOTA methods. On top of that, the synthetic action sequences are tested with different SOTA action recognition approaches in two data augmentation tasks. The experimental results suggest that the proposed method can help boost the action recognition performance with its realistically synthetic data. Since the proposed method has the limitation of generating long-period and consistent action sequences, in the future, the authors will investigate the possibility of extending the proposed work to the long-term action sequence synthesis task and further improving the quality of synthetic action sequences.

The authors have no competing interests to declare that are relevant to the content of this article.

\bibliographystyle{unsrt}
\bibliography{main}
\end{document}